\pdfoutput=1

\documentclass[11pt]{article}

\usepackage[preprint]{acl}
\usepackage{times}
\usepackage{latexsym}

\usepackage[T1]{fontenc}

\usepackage[utf8]{inputenc}

\usepackage{microtype}

\usepackage{graphicx}
\usepackage{booktabs}
\usepackage{xcolor}
\usepackage{subcaption}
\usepackage{algorithm}
\usepackage{algpseudocode}
\usepackage{tcolorbox}
\usepackage{csquotes}
\usepackage{multirow}
\usepackage{hyperref}
\usepackage{amsmath}
\usepackage{longtable}
\usepackage{dsfont}
\usepackage{supertabular}
\tcbuselibrary{skins}

\definecolor{cGrey}{RGB}{248, 248, 248}
\definecolor{cYellow}{RGB}{255,255,3}
\definecolor{cBlue}{RGB}{69,123,157}
\definecolor{cRed}{RGB}{231,56,71}
\definecolor{cRed_1}{RGB}{191,30,46}
\definecolor{cGray}{RGB}{168,218,219}
\definecolor{cBlue_2}{RGB}{5,48,97}
\definecolor{cBlue_1}{RGB}{115,186,214}
\definecolor{cBlue_3}{RGB}{13,76,109}
\definecolor{cBlue_4}{RGB}{64,121,160}
\definecolor{cOrange}{RGB}{250,134,0}
\definecolor{cBlue_6}{RGB}{13,76,109}
\definecolor{cBlue_7}{RGB}{16,106,130}
\definecolor{cBlue_8}{RGB}{19,136,160}
\newcommand*{\affaddr}[1]{#1} 
\newcommand*{\affmark}[1][*]{\textsuperscript{#1}}
\DeclareMathOperator*{\argmax}{arg\,max}

\usepackage{cleveref}
\crefformat{section}{\S#2#1#3}
\crefformat{subsection}{\S#2#1#3}
\crefformat{subsubsection}{\S#2#1#3}
\crefrangeformat{section}{\S#3#1#4 to~\S#5#2#6}
\crefmultiformat{section}{\S#2#1#3}{ and~\S#2#1#3}{, #2#1#3}{ and~#2#1#3}
\Crefformat{figure}{#2Fig.~#1#3}
\Crefmultiformat{figure}{Figs.~#2#1#3}{ and~#2#1#3}{, #2#1#3}{ and~#2#1#3}
\Crefformat{table}{#2Tab.~#1#3}
\Crefmultiformat{table}{Tabs.~#2#1#3}{ and~#2#1#3}{, #2#1#3}{ and~#2#1#3}
\Crefformat{appendix}{#2Appx.~\S#1#3}
\title{\textsc{InferenceDynamics}: Efficient Routing Across LLMs through Structured Capability and Knowledge Profiling}

\author{Haochen Shi\affmark[1]\thanks{\quad Equal Contribution},
Tianshi Zheng\affmark[1]$^{*}$,
Weiqi Wang\affmark[1]$^{*}$,
Baixuan Xu\affmark[1],
Chunyang Li\affmark[1],
Chunkit Chan\affmark[1],
\\
\textbf{
Tao Fan\affmark[1]$^{,}$\affmark[2],
Yangqiu Song\affmark[1],
Qiang Yang\affmark[3]}\\
\affaddr{\affmark[1]The Hong Kong University of Science and Technology}
\affaddr{\affmark[2]WeBank}\\
\affaddr{\affmark[3]The Hong Kong Polytechnic University\\}
\texttt{\{hshiah, tzhengad, wwangbw\}@cse.ust.hk}\\}

\begin{document}
\maketitle
\pagenumbering{arabic}
\begin{abstract}

Large Language Model (LLM) routing is a pivotal technique for navigating a diverse landscape of LLMs, aiming to select the best-performing LLMs tailored to the domains of user queries, while managing computational resources. 
However, current routing approaches often face limitations in scalability when dealing with a large pool of specialized LLMs, or in their adaptability to extending model scope and evolving capability domains. 
To overcome those challenges, we propose \textbf{InferenceDynamics}, a flexible and scalable multi-dimensional routing framework by modeling the capability and knowledge of models.
We operate it on our comprehensive dataset \textbf{RouteMix}, and demonstrate its effectiveness and generalizability in group-level routing using modern benchmarks including MMLU-Pro, GPQA, BigGenBench, and LiveBench, showcasing its ability to identify and leverage top-performing models for given tasks, leading to superior outcomes with efficient resource utilization. The broader adoption of Inference Dynamics can empower users to harness the full specialized potential of the LLM ecosystem, and our code will be made publicly available to encourage further research.







\end{abstract}

\section{Introduction}
The rapid proliferation of Large Language Models (LLMs) has unveiled a rich landscape of specialized capabilities, with different models demonstrating unique strengths across a multitude of domains and tasks~\cite{differentcapabilities1, differentcapabilities2}. This specialization necessitates a sophisticated approach to model selection, where the primary goal is to identify and utilize the LLM best suited to the specific demands of a user's query. LLM routing~\cite{multillm1} emerges as a critical paradigm to address this, creating mechanisms to strategically dispatch queries to the most capable model from a diverse pool, thereby maximizing performance, relevance, and the quality of outcomes, while also considering factors like inference cost and latency.

\begin{figure}[t]
    \centering
    \includegraphics[width=1\linewidth]{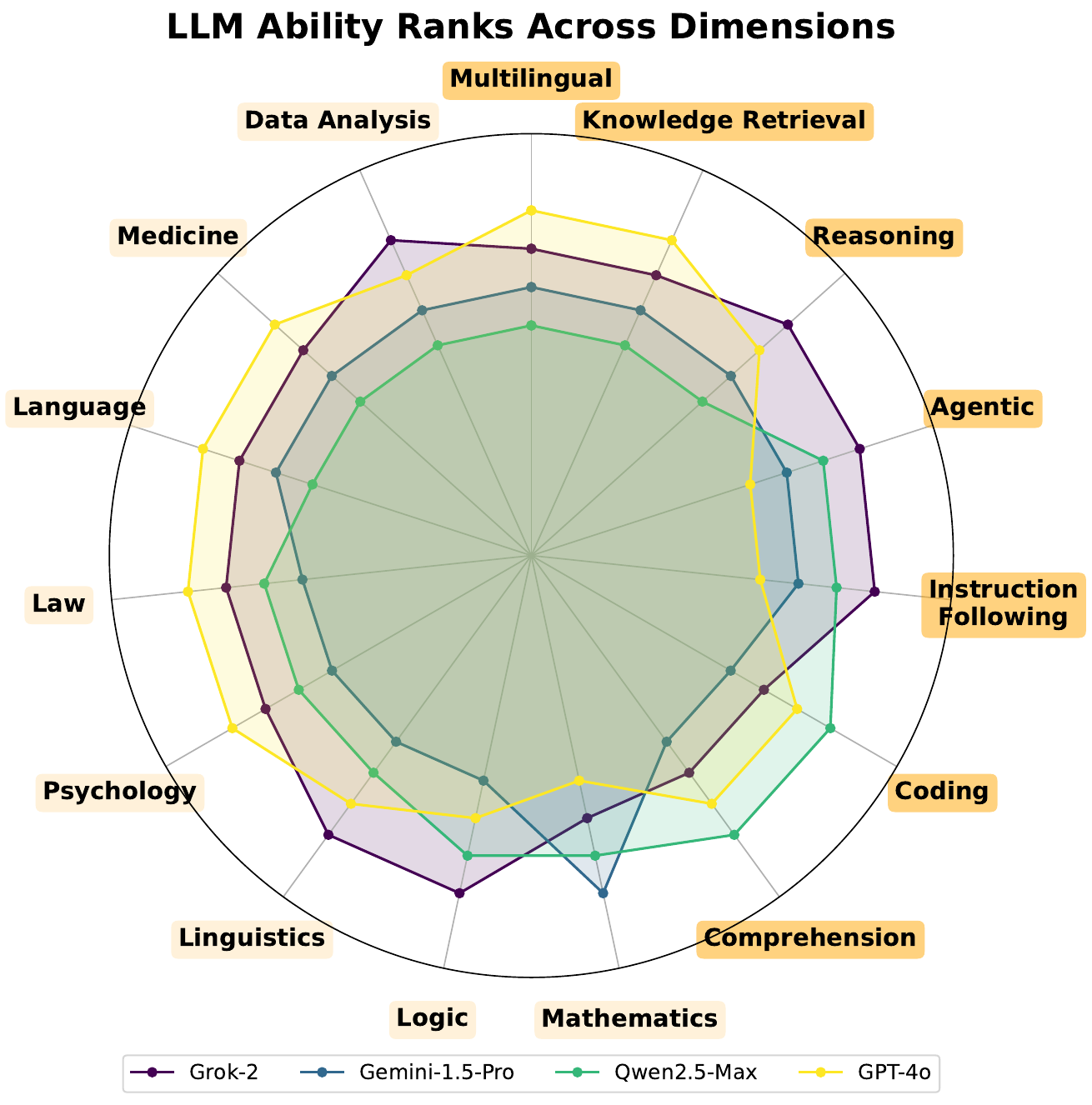}
    \caption{Quantification of Knowledge and Capability of top 4 models among candidate LLMs. }
    \label{fig:llm_rader}
\end{figure}

Early explorations in LLM routing often simplified the selection problem, for instance, by framing it as a binary classification task—e.g., choosing between a generalist small model and a powerful large model. Methods like AutoMix \cite{DBLP:conf/nips/AggarwalMAPMZGR24}, HybridLLM \cite{DBLP:conf/iclr/DingM0SMRLA24}, and RouteLLM \cite{DBLP:conf/iclr/OngAWC0GKS25} demonstrated the viability of this approach, typically focusing on cost-performance trade-offs. While valuable for two-model scenarios, such binary frameworks face inherent scalability challenges, as selecting the best-performing model from many candidates using only pairwise comparisons becomes computationally costly and inefficient.


More recent works have advanced the field by leveraging richer model representations to better evaluate and route LLMs based on their specific capabilities. 
While methods including RouterDC \cite{routerDC}, C2MAB-V \cite{c2mab}, and P2L \cite{p2l} offer more sophisticated mechanisms for capturing model strengths, their primary limitation lies in the significant retraining or recalibration required to effectively support newly introduced LLMs, hindering their agility in a rapidly evolving model landscape. 
Model-SAT \cite{modelsat} aimed to resolve this weakness through human-defined, model-independent capability decompositions. However, its reliance on predefined capability sets undermined adaptability to new capability dimensions, thereby limiting nuanced performance capture in specialized domains.

To address this gap, we introduce \textbf{InferenceDynamics}, a novel system designed for performant, scalable, and adaptable LLM routing. 
\textbf{InferenceDynamics} operates by extracting capability requirements and domain-specific knowledge from incoming queries, modeling the corresponding capabilities and knowledge profiles of available LLMs, and then intelligently routing queries to the most suitable models. 
To demonstrate the effectiveness and generalizability of our approach, we constructed a comprehensive dataset aggregated from 24 diverse benchmarks. 
We then evaluated our routing algorithm on four challenging out-of-distribution (OOD) benchmarks: MMLU-Pro~\cite{MMLUPro}, GPQA~\cite{GPQA}, BigGenBench~\cite{biggen}, and LiveBench~\cite{livebench}. 
Experimental results show that our routing algorithm achieved the highest average score, surpassing the top-performing single LLM by a substantial margin of 1.28 points under optimal routing conditions. 
Furthermore, when operating under cost constraints, our algorithm delivered competitive performance comparable to the best single LLM, while utilizing nearly half the budget.

The contributions of our work are summarized as follows:
\begin{itemize}
    \item We introduce \textbf{RouteMix}, a comprehensive dataset aggregated from 24 diverse benchmarks, specifically curated for rigorously evaluating the generalization capabilities of LLM routing algorithms.
    \item We propose \textbf{InferenceDynamics}, an efficient routing algorithm demonstrating generalization capabilities on previously unseen queries.
    \item Experimental results validate that \textbf{InferenceDynamics} significantly enhances LLM routing, substantially outperforming the leading single model while concurrently reducing computational overhead.
\end{itemize}





\section{Related Works}
\subsection{Multi-LLM System}
A Multi-LLM system~\cite{multillm1} refers to the architecture that combines LLMs to collaboratively solve tasks more effectively than any single model. 
The rapid proliferation of diverse LLMs has spurred significant interest in such systems, which are realized through several architectural patterns.
LLM ensembling~\cite{llmblender, ensembling1} enhances accuracy or robustness by processing the same input through several models and then aggregating their responses. 
Cascaded systems~\cite{cascade1, cascade2, frugalgpt} strategically employ a sequence of models—often initiating with smaller, faster LLMs for initial processing or simpler queries and escalating to more powerful, resource-intensive ones only when necessary—thereby optimizing resource use. 
Furthermore, the development of collaborative LLM agents~\cite{multiagentzihao,multiagent,multiagent2} involves multiple LLMs, with distinct roles or access to different tools, interacting to address complex, multi-step problems that demand sophisticated coordination.
While these multi-LLM approaches demonstrate considerable advancements, they often necessitate querying multiple models, which can increase computational cost and latency. 
Moreover, as the number and diversity of available LLMs continue to grow, it becomes critical to route queries to the most suitable model, effectively balancing performance with operational costs.

\subsection{LLM Routing}
LLM routing seeks to identify the most suitable language model for a given query, with various strategies proposed. Early methods include LLM-Blender~\cite{llmblender}, which employs an ensemble framework querying multiple LLMs to select the optimal response, and AutoMix~\cite{DBLP:conf/nips/AggarwalMAPMZGR24}, which utilizes a smaller model for self-verification before potentially escalating to a larger model. While these can improve performance, their reliance on multiple querying inherently increases latency.
Other strategies, such as HybridLLM~\cite{DBLP:conf/iclr/DingM0SMRLA24} and RouteLLM~\cite{DBLP:conf/iclr/OngAWC0GKS25}, focus on training a binary classifier to choose between a human-defined strong and weak model. However, these methods' efficacy is highly contingent on the subjective definition of model strength and can be computationally expensive when applied to a large pool of LLMs.
More recent research has shifted towards multi-LLM routing. RouterDC~\cite{routerDC},C2MAB-V~\cite{c2mab}, and Prompt-to-Leaderboard~\cite{p2l} trains a parametric router to route queries. Concurrently, ModelSpider~\cite{modelspider} and EmbedLLM~\cite{embedllm} encode LLMs into learnable representations to facilitate routing. Despite these advancements, a significant limitation is the need to retrain the entire routing mechanism when new models are introduced.
Addressing this, Model-SAT~\cite{modelsat} aimed to resolve the retraining weakness through human-defined, model-independent capability decompositions. However, its reliance on predefined capability sets undermined adaptability to new capability dimensions.

\begin{figure*}[t]
\begin{center}
\includegraphics[clip,width=\linewidth]{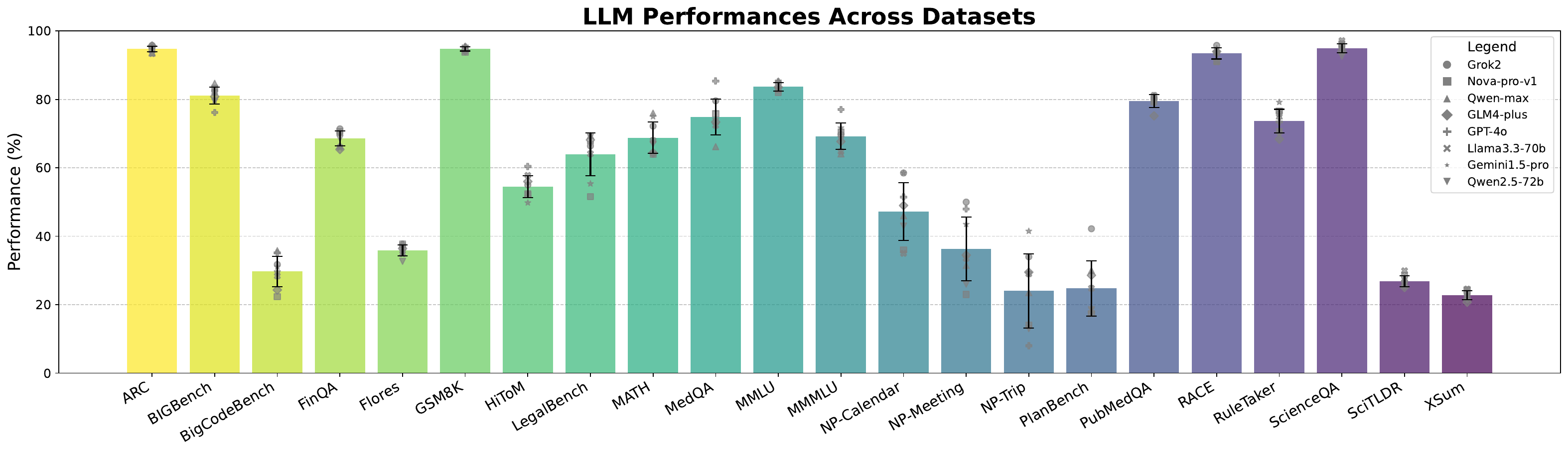}
\end{center}
\vspace{-0.2cm}
\caption{LLM performances across 20 datasets in \textbf{RouteMix}. Dataset labels including "\textit{PlanBench}" indicate subsets of the PlanBench benchmark. For detailed metric information, refer to \Cref{tab:index_set}.}
\label{fig:id_performance}
\vspace{-0.2cm}
\end{figure*}
\section{Methodology}
In this section, we introduce \textbf{InferenceDynamics}, which involves: (i) identifying the knowledge and capability required for a given query, (ii) quantifying the knowledge and capability of LLMs, and (iii) routing queries to LLMs based on their scores.

\subsection{Problem Setup}
Let $\mathcal{M}_T = \{M_1, M_2, \dots, M_t\}$ denote a set of LLMs, and let $\mathcal{D}= \{(\boldsymbol{x}_i, y_i)\}_n$ be a dataset where $\boldsymbol{x}_i$ represents a query and $y_i$ its corresponding ground truth. 
For an unseen query $\boldsymbol{x} \in \mathcal{Q}$, where $x \notin \mathcal{D}$, LLM routing is formalized as a function $\mathcal{R}: \mathcal{Q} \rightarrow \mathcal{M}_T$. 
This function maps the query $x$ to the model $M_{\text{best}} \in \mathcal{M}_T$ that is considered most suitable, based on a joint assessment of both cost and performance.
Our objective is to develop a routing algorithm with the dataset $\mathcal{D}$, that effectively generalizes to OOD queries.

\subsection{Knowledge and Capability Generation}
It is widely acknowledged that no single LLM demonstrates universal proficiency across the full spectrum of query types.
Previous research~\cite{retask, knowledgeboundary} substantiates that distinct queries necessitate specific underlying capabilities and domain-specific knowledge.  
Accordingly, assessing an LLM's aptitude for a given query necessitates identifying the requisite capabilities and knowledge pertinent to that query.
Let $\mathcal{C}$ denote the set of defined LLM capabilities and $\mathcal{K}$ represent the world knowledge space. 
For a given query $x$, we utilize an auxiliary LLM $\mathcal{M} \notin \mathcal{M}_T$ to predict two sets: 
$\mathcal{C}_x = \{c_1, c_2, \dots \mid c_i \in \mathcal{C}\}$: This set comprises the capabilities deemed necessary to address query $x$, ranked in descending order of importance.
$\mathcal{K}_x = \{k_1, k_2, \dots \mid k_i \in \mathcal{K}\}$: This set encompasses the knowledge areas considered essential for resolving query $x$, also ranked in descending order of importance.

Following \citet{llmsurvey}, we categorize capabilities into reasoning, comprehension, instruction following, agentic, knowledge retrieval, coding, and multilingual. 
With regard to the knowledge dimension ($\mathcal{K}_x$), we impose no predefined constraints to fully accommodate its inherent diversity.

\subsection{Scoring}
To quantify the proficiency of a model $M_t$ with respect to specific capabilities and knowledge, we utilize the accessible set $\mathcal{D}$.
The performance score $s^t_i$ of model $M_t$ for a given query-response pair $(\boldsymbol{x_i}, y_i)\in \mathcal{D}_{index}$ is determined by averaging over $K$ independent trials:

\begin{equation*}
s^t_i = \frac{1}{K}\sum_{k=1}^{K}{eval(M_t(\boldsymbol{x_i})_k, y_i)}
\end{equation*}
where $M_t(\boldsymbol{x_i})_k$ is the model's $k$-th generated response to the input query $\boldsymbol{x_i}$, and $eval(\cdot, \cdot)$ represents the query-specific evaluation metric employed to compare the model's response against the ground truth $y_i$.
To incorporate the trade-off between performance and computational expenditure, we record the average computational cost $c^t_i$ incurred by model $M_t$ when processing query $\boldsymbol{x_i}$.

Subsequent to the identification of the knowledge and capability sets and computing the scores for all queries in the set $\mathcal{D}$, we define a refined score for model $M_t$.
This score, $S^\alpha_\beta(M_t, \boldsymbol{x_i}, e)$, quantifies the model's effectiveness for a specific element $e$ (which can be a knowledge item $k \in \mathcal{K}_{\boldsymbol{x_i}}$ or a capability $c \in \mathcal{C}_{\boldsymbol{x_i}}$) associated with query $\boldsymbol{x_i}$. 
Illustrating with a knowledge element $k$, this score is formulated as:


\vspace{-0.5em}
\begin{equation*}
\small
    S^\alpha_\beta(M_t, \boldsymbol{x_i}, k) = {\sum_{j=1}^{|\mathcal{K}_{\boldsymbol{x_i}}|}{(s^t_i-\beta c^t_i) \mathds{1}(k=k_j)\frac{\alpha^{j-1}}{\sum_{m=1}^{|\mathcal{K}_{\boldsymbol{x_i}}|}\alpha^{m-1}}}}
\end{equation*}
\vspace{-1em}

In this formulation, the hyperparameter $\alpha$ serves to attenuate the influence of less critical knowledge elements, based on their rank $j$. The hyperparameter $\beta$ acts as a coefficient penalizing higher computational costs.
The denominator, $\sum_{k=1}^{|\mathcal{K}_{\boldsymbol{x}_i}|} \alpha^{k-1}$, functions as a normalization factor, ensuring that each query contributes equitably to the knowledge score, regardless of the number of knowledge elements it encompasses.

Building upon these per-query, per-element scores, the aggregate score of model $M_t$ for a specific knowledge element $k$ across the entire indexing dataset $\mathcal{D}$ is computed as:
\begin{equation*}
\small
    S^\alpha_\beta(\mathcal{M}_t, \mathcal{D}, k) = \frac{1}{|\mathcal{D}^k|}\sum_{i=1}^{N}S^\alpha_\beta(M_t, \boldsymbol{x_i}, k)
\end{equation*}

where $\mathcal{D}^k = \{(\boldsymbol{x}_i, y_i) \mid k \in \mathcal{K}_i\}$ denotes the subset of query-response pairs in which knowledge $k$ is present in the knowledge set. 
A similar methodology is employed for the computation of aggregate capability scores.

\begin{table*}[th]
\renewcommand\arraystretch{1}
\small
\centering
\setlength{\tabcolsep}{9pt}
\begin{tabular}{cl|cccc|c}
\toprule
& \textbf{Method} & MMLU-Pro & GPQA &BigGenBench& LiveBench & Avg.\\
\midrule
\multicolumn{6}{@{}l}{\textbf{Single Large Language Model}} \\
&Gemini-1.5-Pro &\textbf{82.83} &75.76 &80.92 &\underline{53.79} & 73.33\\
&GPT-4o &79.71 &74.24 &\textbf{85.36} &49.62 & 72.23\\
&Grok-2 &80.14 &76.26 &83.66 &53.26 & 73.33\\
&Qwen2.5-Max &75.86 &71.21 &82.48 &52.77 & 70.58\\
&GLM-4-Plus &79.06 &75.76 &83.27 &47.32 & 71.35\\
&Nova-Pro &77.49 &70.20 &83.01 &44.38 & 68.77\\
&Llama-3.3-70B-Instruct &76.27 &69.70&78.17 &50.67 & 68.70\\
&Qwen-2.5-72B-Instruct &75.41 &73.23 &82.61 &49.83 & 70.27\\
\midrule
& Random &78.26 &72.22 &82.61 & 48.83 & 70.48\\
\midrule
\multicolumn{6}{@{}l}{\textbf{Routing Algorithm (Ours)}} \\
& Routing by \textit{Knowledge} &\underline{80.99} &\textbf{78.28} &82.61 &53.17 & \underline{73.76} \\
& Routing by \textit{Capability} &80.09 &76.26 & 84.18 &53.65 & 73.55\\
& \textit{Inference Dynamics} &80.85 &\underline{77.78} &\underline{84.31} &\textbf{55.57} & \textbf{74.55} \\
\bottomrule
\end{tabular}
\vspace{-0.05in}
\caption{
LLM routing results across four benchmarks are presented.
The metrics we used are introduced in ~\Cref{exp:experiment_seting}.
The best performances are \textbf{bold-faced}, while the second-best performances are \underline{underlined}.  
"Routing by Knowledge" denotes routing decisions made solely based on the knowledge score, whereas "Routing by Capability" refers to routing based only on the capability score.  
"Mixed Routing" indicates a simultaneous consideration of both scores during the routing process.
}
\label{table:main_exp_results}
\vspace{-0.15in}
\end{table*}


\subsection{Routing when inference}
For an unseen query $\boldsymbol{x}$ with its knowledge and capability sets, we compute the knowledge score $KS$ and capability score $CS$ for each candidate model $M_t$ to guide routing. 
The knowledge score is given by:
\begin{equation}
\small
    KS^{\alpha}(M_t,\boldsymbol{x}) = \sum_{i=1}^{|\mathcal{K}_{\boldsymbol{x}}|}S^{\alpha}_{\beta}(M_t, \mathcal{D}, k_i)\frac{\alpha^{i-1}}{\sum_{m=1}^{|\mathcal{K}_{\boldsymbol{x}}|}\alpha^{m-1}},
\end{equation}
The capability score, $CS^{\alpha}(M_t,\boldsymbol{x})$, is computed analogously.
Normalization across both knowledge and capability score calculations ensures that these two distinct types of scores are on a comparable scale, facilitating a balanced routing decision.


The final routing decision is determined by the following algorithm:
\begin{equation}
\small
    \mathcal{R}_{\mathcal{M}_T}(\boldsymbol{x}) = \argmax_{M_t\in\mathcal{M}_T}(\gamma KS^{\alpha}(M_t,\boldsymbol{x})+\delta CS^{\alpha}(M_t,\boldsymbol{x}))
\end{equation}
which aims to identify the model with the highest weighted average of the knowledge and capability scores. 
A key advantage of this framework is its adaptability. New LLMs are efficiently integrated by evaluating them on $\mathcal{D}$ to quantify their knowledge and capability scores, which are then used in routing. Similarly, when queries introduce novel knowledge, the LLMs' scores for this new knowledge can be computed and integrated, refining subsequent routing decisions.

\section{Experiment}
\subsection{Dataset}
In this section, we introduce our comprehensive dataset: \textbf{RouteMix}, which consist of the Index Set and Evaluation Set.

\subsubsection{Index Set}
The term 'Index Set' designates the dataset utilized during the development of our routing algorithm. Given that our methodology is parameter-free, this nomenclature serves to differentiate it from datasets conventionally used in training-dependent methods. The 'Index Set' is thus employed primarily for characterizing and indexing the capabilities and knowledge of LLMs.
To construct a sufficiently diverse 'Index Set' for robust LLM profiling, we have curated 20 distinct datasets. These datasets span a wide array of domains and are instrumental in quantifying the specific knowledge and capabilities of each model.
Comprehensive details regarding the statistics, data processing methodologies, and evaluation metrics for each dataset are presented in ~\Cref{tab:index_set}.

\subsubsection{Evaluation Set}

We incorporate four benchmarks that comprehensively evaluate the LLM as the evaluation set of \textbf{RouteMix}:
(i) MMLU-Pro~\cite{MMLUPro} spans 14 diverse domains and includes approximately 12,000 instances.
(ii) GPQA~\cite{GPQA} consists of multiple choice questions at the graduate level in subdomains of physics, chemistry, and biology. For our evaluation, we utilize the Diamond subset.
(iii) BigGenBench~\cite{biggen} comprises 77 distinct tasks evaluating core abilities of LLM, with a total of 765 human-written instances.
(iv) LiveBench~\cite{livebench} is a real-time updated benchmark with 18 tasks across 6 categories, including math, reasoning, coding, data analysis, language and instruction following. 
In the evaluation, we utilize the snapshot released on 2024-11-25.

\subsection{Experiment Setup}\label{exp:experiment_seting}
For the candidate models, we select eight high-performing LLMs: Gemini-1.5-Pro\cite{gemini1.5}, GPT-4o~\cite{gpt-4o}, Grok-2, Qwen2.5-Max~\cite{qwen2.5}, GLM-4-Plus~\cite{glm-4}, Nova-Pro~\cite{nova}, Llama-3.3-70B-Instruct~\cite{llama3}, and Qwen-2.5-72B-Instruct~\cite{qwen2.5}.
To ensure a fair comparison when testing these models, all parameters and the input prompt are kept consistent across evaluations.
To derive the Knowledge and Capability attributes, we employ GPT-4o-mini to generate these characteristics. 
Since knowledge may include semantically similar phrases, we utilize MiniLM-L6~\cite{minilm} to consolidate Knowledge entries with a cosine similarity score greater than 0.6. 
Additionally, Knowledge entries with a frequency lower than 10 are filtered out and designated as 'Other' knowledge. When the system encounters a query containing previously unseen knowledge elements, these are also classified as 'Other' knowledge.
By default, for unconstrained routing, the parameters $\alpha$ and $\beta$ are set to 0.5 and 0, respectively. 
The weights for the Knowledge and Capability scores are both set to 1.0 by default.
In terms of evaluation, the exact match score is employed for both the MMLU-Pro and GPQA datasets. For BigGenBench, we follow the methodology proposed by \citet{tocotornot}, using GPT-4o-mini as a language model-based judge. 
Instances receiving a score greater than 4 are classified as correct. 
For LiveBench, we adhere to the original evaluation script, and the metric is average score across six categories.
\vspace{0.1cm}
\subsection{Capability and Knowledge Quantification}
The performance of the candidate models on the Index Set is presented in ~\Cref{fig:id_performance}. Generally, these models do not exhibit substantial performance distinctions when evaluated across the entire Index Set. However, their relative strengths become apparent on specific subsets, where different models tend to outperform one another. This observation suggests that the model pool consists of LLMs with broadly comparable overall abilities, yet with varying specializations.\\
\vspace{0.2cm}

Subsequent to the computation of average performance scores, the top four models are selected for more detailed analysis. Their respective capability and knowledge scores are visualized in ~\Cref{fig:llm_rader}. For clarity and simplification in this visualization, we focus on the eight most frequently occurring knowledge elements and capabilities within the Index Set.
The fact that the highest-scoring model changes with the specific knowledge or capability further substantiates the premise: LLMs, even those exhibiting similar aggregate performance levels, possess distinct areas of specialized expertise.

\subsection{Optimal Routing}
The optimal routing results, presented in \Cref{table:main_exp_results}, highlight the clear superiority of our proposed routing strategies. Among these, our \textit{Mixed Routing} strategy, which combines both \textit{Knowledge} and \textit{Capability} scores, achieves the highest average performance, outperforming the best single model, Gemini-1.5-Pro, by a margin of 1.28.
This strategy secures top results on LiveBench and ranks second on GPQA and BigGenBench, demonstrating the effectiveness and versatility of our comprehensive routing algorithm. Additionally, the Routing by \textit{Knowledge} and Routing by \textit{Capability} approaches also deliver strong results, consistently surpassing the best single model and significantly outperforming random routing on average.
Notably, Routing by \textit{Knowledge} excels in knowledge-intensive tasks, achieving the best score on GPQA and the second-best on MMLU-Pro. This underscores its ability to effectively direct queries requiring accurate factual recall and nuanced domain understanding. Similarly, Routing by \textit{Capability} performs exceptionally well on capability-driven benchmarks, particularly on BigGenBench, highlighting the importance of leveraging a model's inherent strengths in complex reasoning and generation tasks. Both approaches play an integral role in the success of the \textit{Mixed Routing} system.

These findings also emphasize that no single LLM universally dominates across all tasks. Models like Gemini-1.5-Pro and GPT-4o exhibit varying strengths, further validating the necessity and advantages of intelligent LLM routing systems.
\subsection{Routing with Constraints}

\begin{figure}[t]
    \centering
    \includegraphics[width=1\linewidth]{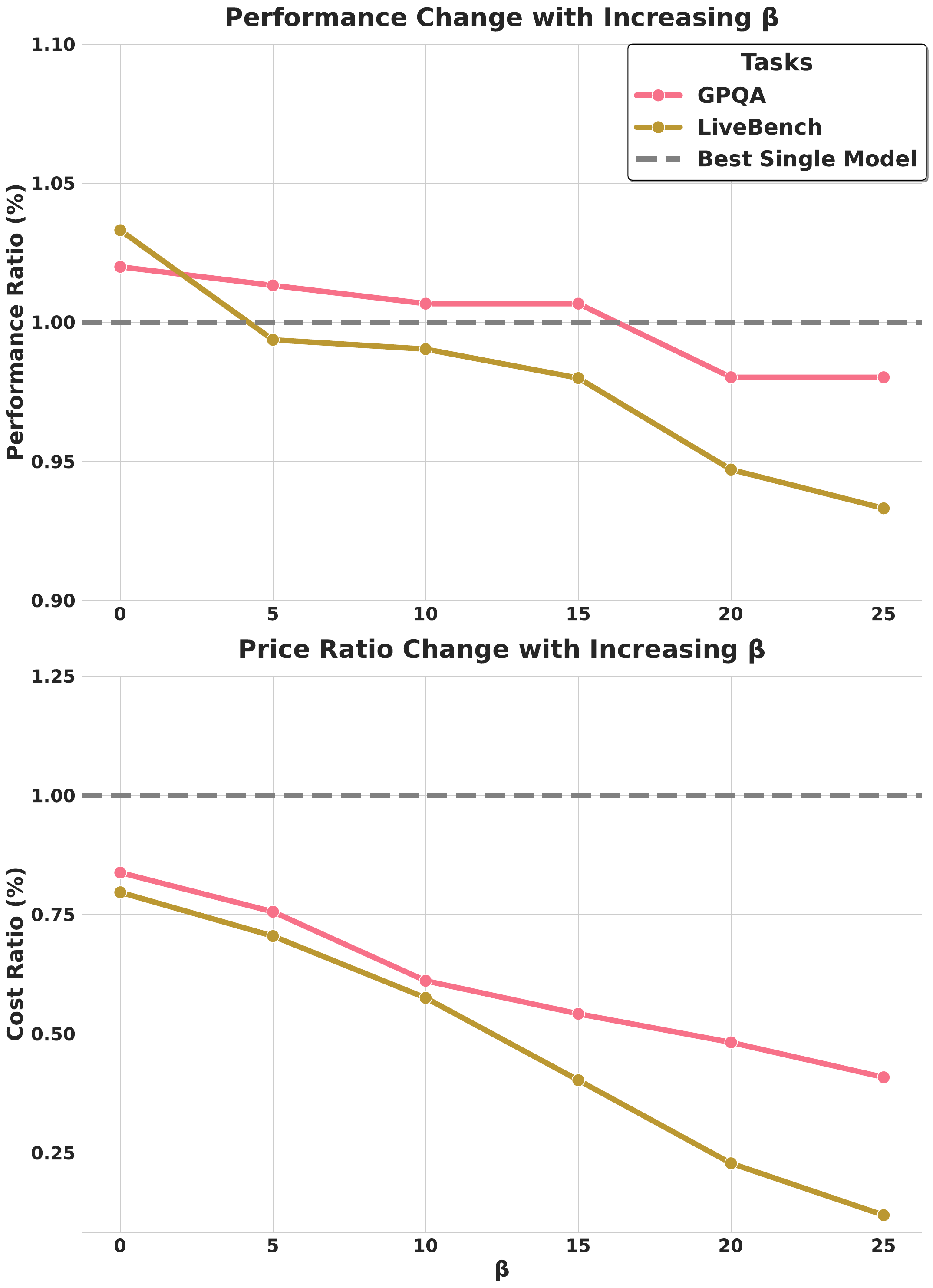}
    \caption{Performance Ratio (\%) and Cost Ratio (\%) variation on GPQA and LiveBench. The "Best Single Model" refers to the most performant LLM for each task.}
    \label{fig:beta_performance}
    \vspace{-0.4cm}
\end{figure}

\begin{figure*}[t]
\begin{center}
\includegraphics[clip,width=\linewidth]{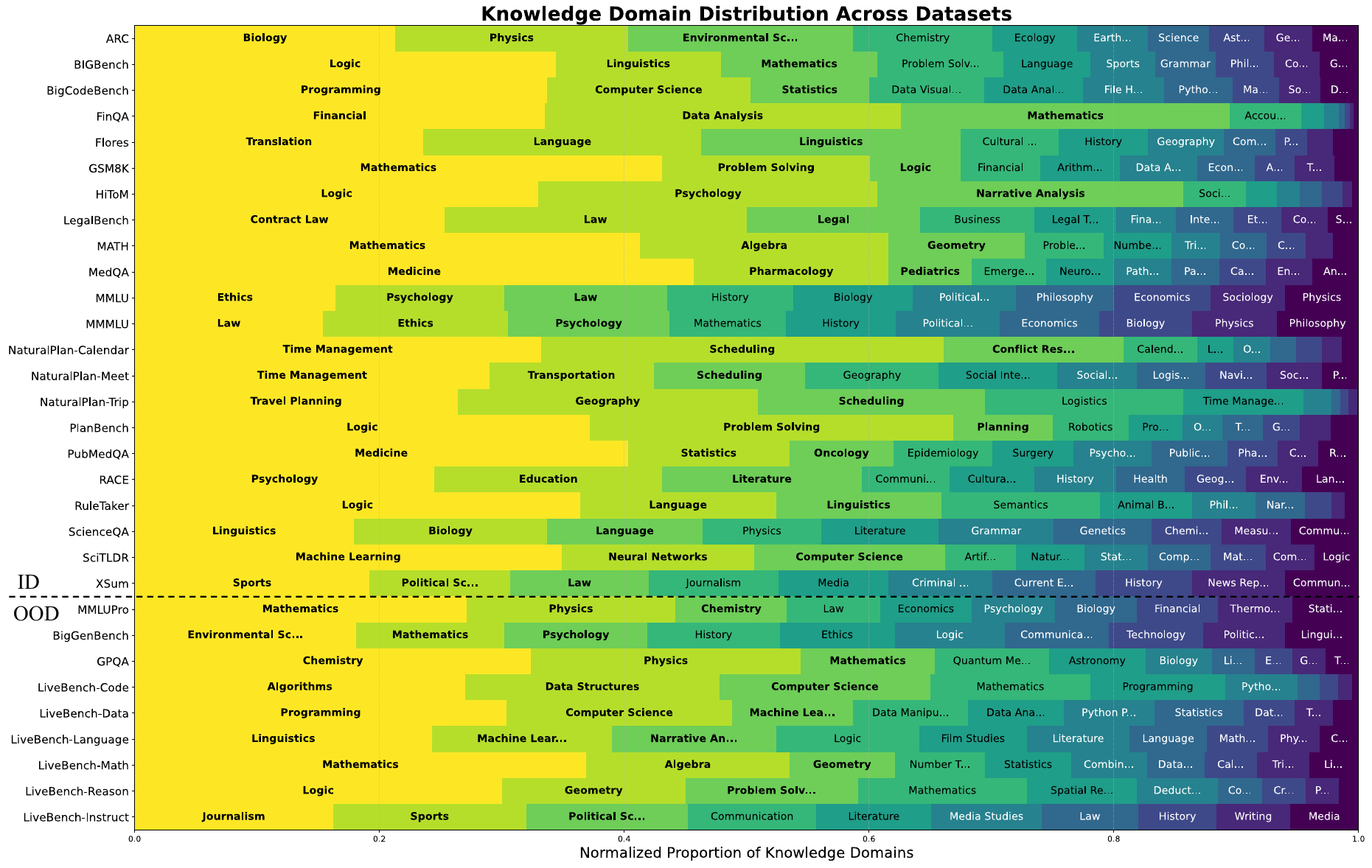}
\end{center}
\caption{Distribution of knowledge domains across 24 datasets in \textbf{RouteMix}. The In-Domain (ID) subset is utilized for quantifying \textit{Knowledge} and \textit{Capability}, while the Out-of-Domain (OOD) subset is employed for evaluating the routing algorithm. Dataset labels including "\textit{LiveBench}" indicate subsets of the LiveBench benchmark, and labels including "\textit{NaturalPlan}" similarly denote subsets of the NaturalPlan benchmark. The algorithm to compute the normalized proportion is included in \Cref{domain_calculation}.}
\label{fig:domain_dist}
\vspace{-0.2cm}
\end{figure*}

\begin{figure*}[t]
\begin{center}
\includegraphics[clip,width=\linewidth]{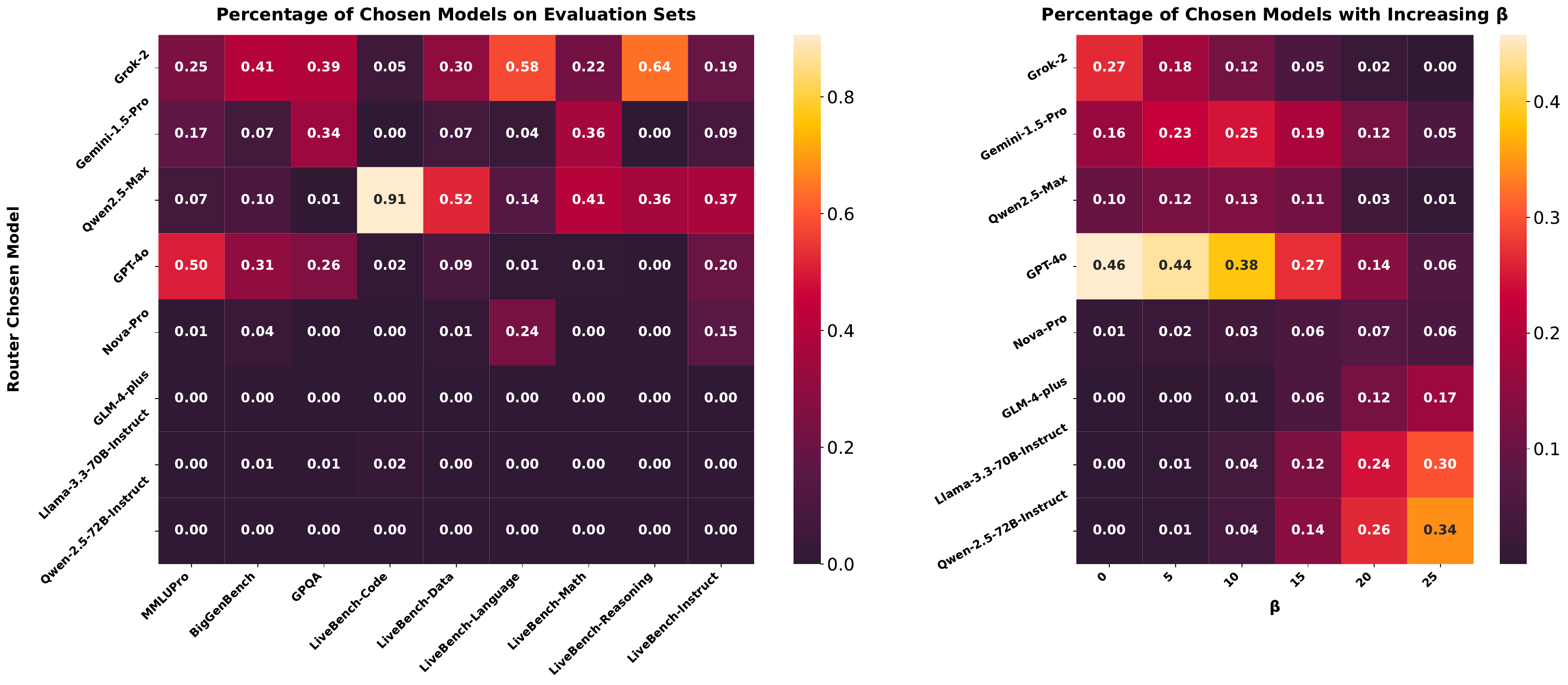}
\end{center}
\caption{Comparative distribution of router-selected models. Lighter colors signify a higher selection ratio for a given model. The left panel details model selection across evaluation benchmarks using the Optimal \textit{Mixed Routing} strategy. The right panel illustrates the impact of an increasing cost penalty coefficient ($\beta$) on the model selection distribution.}
\label{fig:heatmap}
\vspace{-0.2cm}
\end{figure*}

To investigate the system's performance under varying cost constraints, we systematically adjusted the $\beta$ parameter, maintaining all other experimental configurations as previously defined. The evaluation employed two distinct metrics.
The first metric, termed \textbf{Performance Ratio}, quantifies the efficacy of the \textit{Mixed Routing} strategy. 
This is calculated as the ratio of the performance achieved by \textit{Mixed Routing} to that of the best-performing single candidate LLM on the respective benchmark.
The second metric, \textbf{Cost Ratio}, assesses the economic efficiency of the routing algorithm. 
It is defined as the total cost incurred by the routing process (encompassing both knowledge generation and capability assessment costs) relative to the operational cost of the best-performing single LLM.

The empirical results of this sensitivity analysis are depicted in \Cref{fig:beta_performance}. 
In scenarios without stringent price constraints (i.e., $\beta=0$ ), our routing system demonstrates superior performance compared to the best single model, while operating at approximately $80\%$ of the latter's budget. 
As the $\beta$ parameter is incrementally increased, thereby prioritizing cost reduction, the operational cost of the routing algorithm decreases significantly. 
Concurrently, the system maintains a competitive performance level relative to the best single model. 
Notably, at a $\beta$ value of 15, our routing algorithm achieves performance nearly equivalent to the best single model but utilizes only approximately half the associated cost.

An interesting observation is the differential sensitivity of benchmarks to changes in $\beta$. Specifically, the performance and cost metrics for LiveBench, a text generation benchmark, exhibit more pronounced variations in response to adjustments in $\beta$ compared to those observed for GPQA, a question-answering benchmark. This suggests that text generation tasks are more sensitive to the price penalty imposed by $\beta$ than QA tasks.

\section{Analysis}


\subsection{Model Selection}
The distribution of model selections under various conditions is illustrated in \Cref{fig:heatmap}. Consistent with findings in previous works~\cite{routerDC, p2l}, cost-efficient models are infrequently selected in optimal routing scenarios; instead, the strategy predominantly converges towards higher-performing models. For comprehensive benchmarks such as BigGenBench, our approach primarily routes queries to expensive yet high-performing models like GPT-4o and Grok-2, reflecting a tendency to leverage top-tier capabilities for broad-ranging tasks. Conversely, for task sets demanding highly specialized capabilities, the routing algorithm typically assigns queries directly to the most proficient model. For instance, within the coding subset of LiveBench, 91\% of queries are routed to Qwen-Max, which demonstrates the strongest coding capabilities. This model's leading performance in coding is further corroborated by its results on BigCodeBench and its specific Coding capability score, as detailed in \Cref{fig:llm_rader} and \Cref{fig:id_performance}, respectively. These observations collectively indicate that our routing algorithm effectively directs queries to the most suitable models based on specific task demands.

In the context of cost-constrained routing, an increasing cost penalty prompts the router to progressively shift its selections from expensive, top-performing models towards more affordable, albeit less powerful, alternatives.
\subsection{Knowledge Distribution}
As shown in \Cref{fig:domain_dist}, the distribution of generated knowledge highlights the \textit{RouteMix} benchmark's comprehensive span of knowledge domains, ranging from highly specific academic areas to practical applications. 
On datasets with broad knowledge requirements, such as MMLU-Pro, the generated knowledge exhibits a relatively balanced distribution. 
For benchmarks targeting one or two specific domains, like MATH-500, the model typically generates more fine-grained knowledge components related to the core domain. 
This facilitates a more nuanced quantification of the model's domain-specific knowledge.

\subsection{Dynamics Routing}

\begin{figure}[t]
    \centering
    \includegraphics[width=1\linewidth]{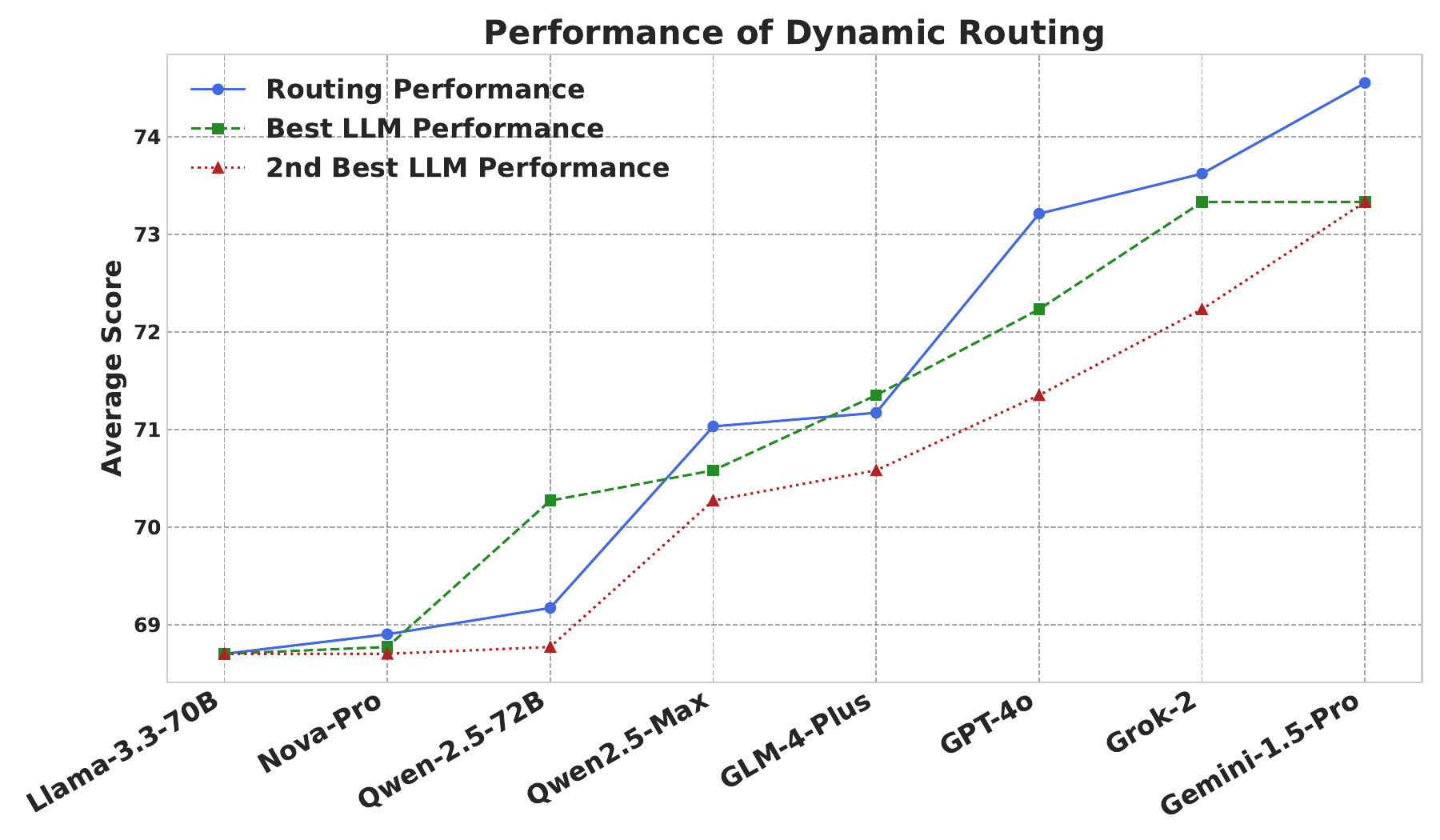}
    \caption{Routing Performance (\%) in Dynamic LLM Pools.}
    \label{fig:dynacmic routing}
\end{figure}
In this section, we investigate the scalability of our framework with respect to dynamic LLM pools. The corresponding results are presented in \Cref{fig:dynacmic routing}. The x-axis in this figure represents the progressive addition of specific new models to the LLM candidate pool. Initially, the pool consists solely of Llama-3.3-70B; subsequently, one new model is added to the candidate pool at each increment along the x-axis.
Notably, our routing algorithm consistently maintains a top-2 performance ranking and surpasses the best single model across the five evaluated candidate pool configurations. This outcome demonstrates the robust scalability of our framework when new models are introduced, crucially without the need for any additional training.

\section{Conclusions}
This paper introduces \textbf{InferenceDynamics}, a scalable and adaptable LLM routing framework that quantifies model capabilities and domain-specific knowledge to match queries with the most suitable LLMs. 
Evaluated on the new comprehensive RouteMix benchmark, InferenceDynamics demonstrated superior performance, outperforming the best single LLM by 1.28 on average and achieving comparable results at approximately half the cost under budget constraints. 
Key contributions include the \textbf{RouteMix} dataset for evaluating generalization and the \textbf{InferenceDynamics} algorithm, which generalizes to unseen queries and effectively routes them within dynamic model pools without retraining. Our work enables more efficient and tailored utilization of the diverse LLM ecosystem.

\newpage

\section*{Limitations}

Despite the promising results and the robust design of InferenceDynamics, several limitations warrant discussion and offer avenues for future research:

 \paragraph{Niche Suitability for Highly Constrained Environments} InferenceDynamics is engineered for scalability and adaptability, demonstrating its strengths when dealing with a large, diverse, and evolving pool of LLMs, or when new capability and knowledge domains are frequently encountered. However, in scenarios characterized by a very limited and static set of LLMs and a narrowly defined, unchanging task scope, a dedicated learning-based routing approach (e.g., a fine-tuned classifier) might be more appropriate or yield marginally superior, hyper-specialized performance. Our framework prioritizes generalizability and efficient adaptation to dynamic conditions, which is a different niche than hyper-optimization for small, fixed-scope problems.

  \paragraph{Benchmark-Driven Evaluation vs. Real-World Application Complexity} The current evaluation of InferenceDynamics relies on the comprehensive RouteMix dataset, which is composed of various established benchmarks. While these benchmarks cover a wide array of tasks and domains, they may not fully capture the intricacies and dynamic nature of real-world application systems. For instance, the utility and performance of InferenceDynamics in more complex, interactive systems like multi-agent environments, where task allocation might depend on evolving collaborative states, have not been explicitly tested. Exploring the deployment and effectiveness of InferenceDynamics in such real-application scenarios remains an important direction for future work.

Addressing these limitations will be crucial for broadening the applicability and enhancing the robustness of InferenceDynamics and similar LLM routing frameworks.

\section*{Ethics Statement}
Our study utilizes publicly available datasets and accesses Large Language Models (LLMs) through their respective APIs. The ethical considerations pertaining to this research are as follows:\\
\textbf{Datasets: } This research exclusively employs publicly available datasets, strictly for academic research purposes. We affirm that no personally identifiable information or private data was involved in our study.\\
\textbf{LLM APIs: } Our application of LLMs via APIs rigorously conforms to the policies set forth by the API providers. This includes adherence to fair use guidelines and respect for intellectual property rights.\\
\textbf{Transparency: } In line with standard academic research practices, we provide detailed descriptions of our methodology and the prompts utilized in our experiments. Furthermore, the source code for this research will be made publicly available upon the acceptance of this paper.
\bibliography{custom}

\newpage


\appendix
\onecolumn
\section{Benchmark Overview Table}\label{tab:index_set}

\begin{longtable}{@{} l p{6cm} p{2cm} p{2cm} @{}} 
\caption{Overview of Benchmarks, Data Processing, Prompts, and Metrics} \\ 
\label{tab:benchmark_overview_detailed_4col} \\ 
\toprule
\textbf{Benchmark Name} & \textbf{Data Processing Manner} & \textbf{Prompt Type} & \textbf{Metric Used} \\ 
\midrule
\endfirsthead 

\multicolumn{4}{c}
{{\tablename\ \thetable{} -- continued from previous page}} \\
\toprule
\textbf{Benchmark Name} & \textbf{Data Processing Manner} & \textbf{Prompt Type} & \textbf{Metric Used} \\ 
\midrule
\endhead 

\bottomrule
\multicolumn{4}{r}{{Continued on next page...}} \\ 
\endfoot 

\bottomrule
\endlastfoot 

ARC~\cite{allenai:arc} & Sample 500 instances according to the portion of ARC-Easy and ARC-Challenge.  & Zero-shot DA & Accuracy \\
BigBench-Hard~\cite{BBH} & Sample 40 instances from each category except $web_of_lies$, to avoid collision with LiveBench. Formulate into MCQA for Yes/No and QA question. Remain the free-response question unchanged. & Zero-shot CoT & Exact Match (EM) \\
BigCodeBench~\cite{zhuo2024bigcodebench} & We directly use the BigCodeBench-Hard subset, with 148 instances. & DA for code completion & Pass@1 \\
FinQA~\cite{finqa} & Sample 500 instances from the dataset. & CoT from \citet{tocotornot} & Exact Match(EM) \\
Flores200~\cite{flores} & We incorporate the top10 commonly used language except for English. And sample 100 instances for each language. & Translation Prompt & Chrf++~\cite{flores} \\
GSM8K~\cite{gsm8k} & Sample 500 instances from the dataset. & CoT from \citet{tocotornot} & Exact Match(EM). \\
HiToM~\cite{hitom} & Sample 500 instances under CoT settings. & CoT from Official Repo & Accuracy \\
LegalBench~\cite{legalbench} & Sample 4 instances from each category except for short answering task, resulting in 616 instances. & Few-shot DA & Accuracy \\
MATH~\cite{math} & We use the subset MATH-500. & CoT from \citet{tocotornot} & Exact Match(EM) \\
MedQA~\cite{medqa} & Sample 500 instances from the dataset & DA & Accuracy \\
MMLU~\cite{mmlu}& We sample instances according to the portion of different categories, and make sure each category has at least 10 instances. Resulting in 1262 instances. & DA & Accuracy \\
MMMLU~\cite{mmlu} & We sample 100 instances for all languages except for English. Result in 1400 instances. & DA & Accuracy \\
NaturalPlan~\cite{naturalplan} & Sample 200 instances from each subset, including scheduling, calendar meeting, and trip planning.& DA& Accuracy \\
PlanBench~\cite{planbench} & Use the subset of PlanGeneration in BlocksWorld. &DA &Accuracy \\ 
PubMedQA~\cite{pubmedqa} & Sample 500 instances from original dataset & DA & Accuracy \\
RACE~\cite{race} & Sample 500 instances from original dataset & DA & Accuracy \\
RuleTaker~\cite{ruletaker} & Sample 500 instances from original dataset & DA & Accuracy \\
ScienceQA~\cite{scienceqa} & Sample 500 instances which don't have corresponding picture. & DA & Accuracy \\
SciTLDR~\cite{scitldr} & Directly use the test set & Summarization Prompt & RogueL. \\
XSum~\cite{xsum} & Sample 500 instances for the dataset & Summarization Prompt & RogueL \\

\end{longtable}
Specifically, when quantifying the capability and knowledge of LLMs for translation and summarization tasks, we establish a performance threshold. An output is considered correct if its evaluation score or relevant metric exceeds this threshold.

\section{Knowledge Domain Distribution}\label{domain_calculation}

The dataset's knowledge domain distribution is determined by a weighted rank approach. For each domain \( D \in \mathcal{D} \) (where \( \mathcal{D} \) is the set of all unique domains), its frequency at each rank \( r \) (denoted \( F_{D,r} \), for \( r=1, \dots, N \)) is multiplied by a corresponding rank weight \( W_r \) (typically \( W_r = 1/r \)). These products are summed to yield a weighted score \( S_D \):
\[
S_D = \sum_{r=1}^{N} (F_{D,r} \times W_r)
\]
The final distribution percentage \( P_D \) for each domain is then its \( S_D \) normalized by the sum of all domain weighted scores (\( S_{\text{total}} = \sum_{D' \in \mathcal{D}} S_{D'} \)), expressed as a percentage:
\[
P_D = \left( \frac{S_D}{\sum_{D' \in \mathcal{D}} S_{D'}} \right) \times 100\%
\]
This method ensures higher-ranked domain occurrences contribute more significantly, with all \( P_D \) summing to 100\%.


\section{BigGenBench Evaluation}
Following \citet{tocotornot}, we employ GPT-4o-mini as LLM-as-a-Judge to evaluate the BigGenBench, and instances with a score larger than 4 is considered correct.
The specific prompt is shown below:
\begin{tcolorbox}[title={Prompt for evaluation BigGenBench}, colback = cGrey, colframe = black,  coltitle=white,fonttitle=\bfseries\small, center title,fontupper=\small,fontlower=\small]
Task Description:\\
            An instruction (might include an Input inside it), a response to evaluate, a reference answer that gets a score of 5, and a score rubric representing a evaluation criteria are given.\\
            1. Write a detailed feedback that assess the quality of the response strictly based on the given score rubric, not evaluating in general.\\
            2. After writing a feedback, write a score that is an integer between 1 and 5. You should refer to the score rubric.\\
            3. The output format should look as follows: "Feedback: (write a feedback for criteria) [RESULT] (an integer number between 1 and 5)"\\
            4. Please do not generate any other opening, closing, and explanations.\\
            \\
            The instruction to evaluation:\\
            {example question}\\
\\
            Response to evaluate:\\
            {example solution}\\
\\
            Reference Answer (Score 5):\\
            {reference score}\\
\\
            Score Rubrics:\\
            Criteria:\\
            {criteria}\\
\\
            Description of a Score 1 response:\\
            {score1 description}\\
\\
            Description of a Score 2 response:\\
            {score2 description}\\
\\
            Description of a Score 3 response:\\
            {score3 description}\\
\\
            Description of a Score 4 response:\\
            {score4 description}\\
\\
            Description of a Score 5 response:\\
            {score5 description}\\
\\
            Feedback:\\
            Remember, you must strictly evaluate the response based on the given score rubric, and finish your output in the format of "(...) [RESULT] <score>", where <score> is a number between 1 and 5.
\end{tcolorbox}



\section{Prompt of Knowledge and Capability Generation}\label{kcprompt}
The specific prompt for knowledge and capability generation is shown below:
\begin{tcolorbox}[title={Prompt for evaluation BigGenBench}, colback = cGrey, colframe = black,  coltitle=white,fonttitle=\bfseries\small, center title,fontupper=\small,fontlower=\small]
The capabilities of Language Models include the following:\\
\\
        - Reasoning: Ability to logically analyze information, draw conclusions, and make inferences.\\
        - Comprehension (Applicable to queries involving long passage comprehension): Understanding and interpreting the meaning, context, and nuances of extended or complex long-context text, such as lengthy documents, multi-paragraph inputs, or intricate narratives.\\
        - Instruction Following (Applicable to queries involving several constraints): Accurately adhering to explicit user-provided guidelines, constraints, or formatting requirements specified within the query.\\
        - Agentic: Capacity related to agent-like behavior, such as actively formulating plans, strategically deciding steps, and autonomously identifying solutions or actions to achieve specific goals or complex tasks.\\
        - Knowledge Retrieval: Accessing and presenting accurate factual information from pre-existing knowledge.\\
        - Coding: Generating, interpreting, or debugging computer programs and scripts.\\
        - In-context Learning: Learning from examples or context provided within the current interaction without additional training.\\
        - Multilingual (Must rank it in top3 when queries involving languages other than English): Understanding, generating, or translating content accurately across multiple languages.\\
        Given the Query below:\\
\\
        1. Identify and list the *LLM Capabilities* from the definitions above that are directly and significantly required to effectively address the query.\\
        2. Identify and list the general *Knowledge Domains* (e.g., categories, subject areas) most pertinent to solving the problem presented in the query.\\
        List the selected Capabilities first, ranked from most important to least important. Then, list the identified Knowledge Domains, also ranked from most important to least important. *Do not provide any justification or explanation* for your selections or rankings.\\
\\
        Example:\\
        Query: "{Solve the following financial problem efficiently and clearly. Output the final answer as: boxed{answer}. \\
        Where [answer] is just the final number or expression that solves the problem. Keep the answer to five decimal places if it is a number, and do not use percentages; keep the decimal format.\\
        Problem: what is the net change in net revenue during 2016 for Entergy Mississippi, Inc.? the 2015 net revenue of amount (in millions) is 696.3; the 2016 net revenue of amount (in millions) is 705.4; Entergy Mississippi, Inc.}"
        Capabilities: Reasoning, Knowledge retrieval\\
        Knowledge: {
        1. Financial
        2. Math
        3. Data Analysis
        ...
        }
        Query: {input prompt}
\end{tcolorbox}

\end{document}